\documentclass[conference]{IEEEtran}
\IEEEoverridecommandlockouts

\usepackage{cite}
\usepackage{url}
\usepackage{hyperref}
\usepackage{amsmath,amssymb,amsfonts}
\usepackage{algorithm, algpseudocode}
\usepackage{graphicx}
\usepackage{textcomp}
\usepackage{tikz}
\usepackage{pgfplots}
\usepgfplotslibrary{groupplots}  
\usepackage{subcaption}  
\usepackage{xcolor}
\usepackage{bm}
\usepackage{csvsimple}
\usepackage{filecontents} 
\usepgfplotslibrary{units}  
\pgfplotsset{compat=1.18}
\usetikzlibrary{external}
\usepgfplotslibrary{external}
\usepackage{tabularx}
\usepackage{booktabs}

\newcommand{\graph}{\mathcal{G}}
\newcommand{\nodes}{\mathcal{V}}
\newcommand{\cluster}{\mathcal{C}}
\newcommand{\localdset}{\mathcal{D}}
\newcommand{\vx}{\mathbf{x}}
\newcommand{\vX}{\mathbf{X}}
\newcommand{\vmu}{\boldsymbol{\mu}}
\newcommand{\vpi}{\boldsymbol{\pi}}
\newcommand{\vcov}{\boldsymbol{\Sigma}}
\newcommand{\neib}{\Gamma}

\newcommand{\norm}[1]{\left\lVert#1\right\rVert}

\def\BibTeX{{\rm B\kern-.05em{\sc i\kern-.025em b}\kern-.08em
		T\kern-.1667em\lower.7ex\hbox{E}\kern-.125emX}}

\begin{document}
	\title{Graph-Regularized Learning of Gaussian Mixture Models \\
		\thanks{Funded by the Research Council of Finland (Decision \#363624),
		Jane and Aatos Erkko Foundation (Decision \#A835),
		 Business Finland}
	}
	\author{\IEEEauthorblockN{1\textsuperscript{st} Shamsiiat Abdurakhmanova}
	\IEEEauthorblockA{\textit{dept. of Computer Science} \\
		\textit{Aalto University }\\
		Helsinki, Finland \\
		shamsiiat.abdurakhmanova@aalto.fi}
	\and
	\IEEEauthorblockN{2\textsuperscript{nd} Alex Jung}
	\IEEEauthorblockA{\textit{dept. of Computer Science} \\
		\textit{Aalto University }\\
		Helsinki, Finland \\
		alex.jung@aalto.fi}
	}
	
\maketitle
\begin{abstract}
	We present a graph-regularized learning of Gaussian Mixture Models (GMMs) in distributed settings with heterogeneous and limited local data. The method exploits a provided similarity graph to guide parameter sharing among nodes, avoiding the transfer of raw data. The resulting model allows for flexible aggregation of neighbors' parameters and outperforms both centralized and locally trained GMMs in heterogeneous, low-sample regimes.
\end{abstract}

\section{Introduction}
We propose GraphFed-EM, a federated Gaussian Mixture Model in which local nodes collaboratively learn a personalized probabilistic model through graph-based regularization, without exchanging raw data. The algorithm is adapted for decentralized settings, incorporating an aggregation step that promotes parameter similarity among connected nodes.

GraphFed-EM is beneficial for heterogeneous collections of local datasets, where a single global GMM may fail due to dataset dissimilarity and local training is challenging in low-sample regimes (small sample sizes or high-dimensional features). We illustrate this with synthetic clustered datasets, where datasets within each cluster share common GMM parameters but differ across clusters. We also test our algorithm on synthetic and MNIST datasets with skewed feature distributions. Unlike prior federated EM approaches that assume identical GMM parameters across clients, GraphFed-EM flexibly aggregates parameters via a similarity graph while respecting client-specific variations.

The remainder of the paper is organized as follows: Section~\ref{sec:prelim} reviews Gaussian Mixture Models and EM; Section~\ref{sec:graph-em} presents GraphFed-EM; Section~\ref{sec:related} discusses related work; Section~\ref{sec:experiments} reports experiments; Section~\ref{sec:analyses} presents an interpretation of GraphFed-EM as regularized EM; Section~\ref{sec:conclusions} concludes.
	
\section{Preliminaries on Finite Gaussian Mixtures}
\label{sec:prelim}
The Gaussian mixture distribution can be written as a weighted sum of $K$ Gaussians \cite{bishop2006pattern}:
\begin{align}
	p(\vx) = \sum_{k=1}^{K}\pi_k \mathcal{N}(\vx \mid \vmu_k, \vcov_k)
	\label{eq:prob}
\end{align}
where $\pi_k$ is the mixture weight for component $k$, $\vmu_k$ and $\vcov_k$ are the mean vector and covariance matrix of component $k$, and $\mathcal{N}(\vx \mid \vmu_k, \vcov_k)$ is the multivariate Gaussian probability density function.
Given a set of  i.i.d. observations $\vX=(\vx_1^T, \dots, \vx_N^T)$, the data can be modeled using a mixture of Gaussians, with log-likelihood
\begin{align*}
	\text{ln}p(\vX \mid \vpi, \vmu, \vcov) = \sum_{n=1}^{N} \text{ln} \left\{ \sum_{k=1}^{K}\pi_k \mathcal{N}(\vx \mid  \vmu_k, \vcov_k) \right\} 
\end{align*}
The most commonly used method to maximize this function is the expectation-maximization (EM)
algorithm \cite{dempster1977maximum}. Briefly, the E step of the algorithm consists of computing responsibilities $\gamma_{kn}$ for each data point $\vx_n$ and component $k$:
\begin{align*}
\gamma_{kn} = \frac{\pi_k\,\mathcal{N} \left(\vx_n \mid \vmu_k,\vcov_k\right)}{\sum_{j=1}^{K} \pi_j\,\mathcal{N} \left(\vx_n \mid \vmu_j,\vcov_j\right)}.
\end{align*}
 and M step consists of parameters update step according to closed-form solutions:
\begin{align*}
	\pi_k &= \frac{N_k}{N}, \quad \text{with } N_k = \sum_{i=1}^{N} \gamma_{nk} \\
	\vmu_k &= \frac{1}{N_k} \sum_{n=1}^{N} \gamma_{nk} \vx_n \\
	\vcov_k &= \frac{1}{N_k} \sum_{n=1}^{N} \gamma_{nk} (\vx_n - \vmu_k)(\vx_n - \vmu_k)^T
\end{align*}

We initialize the EM algorithm by setting local GMM means to KMeans centroids and estimate covariances and mixture weights from the dataset split induced by the KMeans cluster assignments. To ensure numerical stability during training, we regularize covariance matrices using shrinkage and diagonal loading techniques.

\section{Graph-Regularized EM Algorithm}
\label{sec:graph-em} 
		
Consider a network (graph) $\graph$ of nodes (clients) connected via adjacency matrix $A$. Each node $i$ holds a local dataset $\localdset^{(i)} := \{ \vx_1^{(i)}, \dots, \vx_{N_i}^{(i)}\}$, where each sample $\vx_r^{(i)} \in \mathbb{R}^d$ is drawn from a Gaussian mixture as in Eq.~\ref{eq:prob}. Node $i$ is connected to its neighbors $j \in \Gamma(i)$ via undirected edges weighted by $A_{ij}$. The graph is assumed to encode similarity among the GMM parameters of connected nodes.

GraphFed-EM training consists of $T_i$ local EM updates (Section \ref{sec:prelim}), followed by an aggregation step. Each node $i$ shares its locally estimated GMM parameters  $\boldsymbol{\theta}^{i} = \{\vmu^{i}, \vcov^{i}, \vpi^{i} \}$ with neighbors to compute a weighted average of its own and neighbors' parameters.
Since the ordering of components is arbitrary, we first align the $K$ components using the linear sum assignment solver from \href{https://docs.scipy.org/doc/scipy/reference/generated/scipy.optimize.linear_sum_assignment.html}{SciPy} on a Bhattacharyya distance matrix.
The aligned parameters are aggregated as follows:
\begin{align}
	\label{eq:aggr}
	\boldsymbol{\theta}_{aggr, k}^{i} = \frac{N_{k}^{i} \boldsymbol{\theta}_{k}^{i} + \sum_{j \in \neib(i)} A_{ij} N_{k}^{j} \boldsymbol{\theta}_{k}^{j} }{N_{k}^{i} + \sum_{j \in \neib(i)} A_{ij} N_{k}^{j}}
\end{align}
with mixing coefficients normalized by $\sum_{k=1}^{K} N_{k}^{i} + \sum_{j \in \neib(i)} A_{ij} N_{k}^{j}$ to ensure $\sum_{k=1}^{K} \pi_k^i=1$. In other words, this corresponds to a responsibility-weighted aggregation of GMM parameters.
Furthermore, the degree of parameter aggregation is controlled by $\alpha$:
\begin{align}
	\label{eq:blending}
	\boldsymbol{\theta}_{k}^{i}  \gets  (1 - \alpha) * \boldsymbol{\theta}_{k}^{i}  + \alpha * \boldsymbol{\theta}_{aggr,k}^{i} 
\end{align}
After aggregation, covariance matrices are symmetrized and diagonal-regularized to ensure positive definiteness.
\begin{algorithm}[t]
	\caption{Graph-Regularized EM Algorithm}
	\label{alg:fed-em}
	\begin{algorithmic}[1]
		\State \textbf{Input:} Empirical graph $\graph := \{\nodes, \mathcal{E}\}$ with $N$ nodes and adjacency matrix $A$. Local dataset $\mathcal{D}^{(i)}$ on each node, $|\mathcal{D}^{(i)}| = N_i$ , number of components $K$, iterations $T$, local iterations $T_i$, aggregation strength $\alpha$
		\State \textbf{Initialize:} GMM parameters $\boldsymbol{\theta}^{(i,0)} = (\vpi^{(i,0)}, \vmu^{(i,0)}, \vcov^{(i,0)})$ for each node $i \in \nodes$
		\For{$t = 1$ to $T$}
				\For{each node $i$}
					\For{$t_i = 1$ to $T_i$}
						\State {Compute $\boldsymbol{\theta}^{(i,t_i)}$ with local EM updates}
					\EndFor
					\State {Share local estimates  $\boldsymbol{\theta}^{(i,T_i)}$ with neighbors $\Gamma(i)$}
			\EndFor
			\For{each node $i$}
				\State Update $\boldsymbol{\theta}^{(i,T_i)} \gets \boldsymbol{\theta}^{(i,T_i)}$ via \ref{eq:blending}
			\EndFor
		\EndFor
		\State \Return $\{\boldsymbol{\theta}^{(i,T)}\}_{i \in \nodes}$
	\end{algorithmic}
\end{algorithm}
		
\section{Related Work}
\label{sec:related}

Our GraphFed-EM algorithm is an instance of the regularized EM (see Section~\ref{sec:analyses}). Regularization in EM has been used to improve stability and convergence in low-sample regimes \cite{houdouin2023regularized, yi2015regularized}, and to simplify mixture models with redundant components \cite{li2005regularized}.

With growing data volumes and device constraints, distributed EM has become necessary. Most approaches adopt a client–server setup to train and share a single global model \cite{zhang2022distributed}, typically assuming i.i.d. data. Some works address data heterogeneity: for example, FedGenGMM \cite{pettersson2025federated} aggregates client-trained GMMs by re-weighting components, generating synthetic data, training a global model, and redistributing it. \cite{dieuleveut2111federated} addresses the estimation of shared global GMM parameters, while introducing communication-efficient EM for federated learning with non-i.i.d. data and partial participation. Closely related to our approach, FedGrEM \cite{tian2023towards} employs a gradient-based EM algorithm with a regularization term that penalizes deviations from the central aggregate parameters.

\section{Numerical Experiments}
\label{sec:experiments}

We evaluate GraphFed-EM on synthetic and MNIST datasets modeling common client heterogeneity. The synthetic datasets include: (i) clustered GMMs with parameters $\{\boldsymbol{\theta}^{(c)}\}_{c=1}^{C}$, where datasets sharing the same parameters form a cluster $\cluster^{(c)}$, and (ii) a shared mixture with node-specific mixing coefficients $\vpi^i$. MNIST is used to create skewed label distributions across clients.
Performance is evaluated on local validation sets (sample size $N_i^{(val)}=500$) by average log-likelihood and normalized mutual information (NMI) \cite{strehl2002cluster} to quantify clustering accuracy. We also report centroid estimation error:
\begin{align}
	\mu_{err} = \frac{1}{NK} \sum_{i \in \nodes} \sum_{k=1}^{K} \norm{\vmu_k^{c_i} - \vmu_k^i}_2^2
\end{align}
where $\vmu_k^{c_i}$ denotes the true cluster centroid, used to generate the data for node $i$. Components are matched using the Bhattacharyya distance (Section \ref{sec:graph-em}).

All experiments are repeated 10 times, with mean and standard errors shown as y-error bars or shaded regions. Unless stated otherwise, GraphFed-EM uses aggregation strength $\alpha=1$, $T=10$ iterations, and $T_i=5$ local EM steps.
Code for these and additional experiments can be found on \href{https://github.com/shamPJ/Graph-GMM}{GitHub}.

\subsection{Synthetic Dataset - Clustered setting}
\label{sec_exp_clustered_ds}

We test GraphFed-EM on a synthetic dataset represented by a graph $\graph$ with $N$ nodes partitioned into $C$ equal-sized clusters $\{\cluster^{(c)}\}_{c=1}^{C}$, where node $i$ belongs to a cluster $\cluster^{(c_i)}$. Edges are independent Bernoulli variables $b_{ij} \in \{0,1\}$, present with probability $p_{in}$ if $i$ and $j$ share a cluster and $p_{out}$ otherwise; edges are unweighted, so $A_{ij}=1$ when $b_{ij}=1$. Each node $i$ holds local data $\localdset^{(i)}=\{\vx^{(i,1)},\dots,\vx^{(i,N_i)}\}$ sampled from a $K$-component GMM with parameters $(\mu_k^{(c_i)},\Sigma_k^{(c_i)},\pi_k^{(c_i)})$ shared among nodes in the same cluster. To introduce heterogeneity, the means and covariance matrices undergo random rotation and shear transformations applied consistently across clusters.

\textbf{\textit{When does pooling local datasets make sense?}}
We compare local training (each local dataset independently), centralized training (all datasets combined), and GraphFed-EM with either direct cluster-wise pooling or an aggregation step (Algorithm~\ref{alg:fed-em}); the latter two, given true cluster identities, are termed Oracles. To isolate the benefit of aggregation over pooling, we also trained GraphFed-EM on cluster-wise pooled datasets.

Experiments used $C=5$ clusters with $N=25$ nodes (5 per cluster), $K=3$ components, feature dimension $d \in \{2,6,10\}$, and local training sizes $N_i \in \{10,50,100\}$. The connectivity matrix was set with $p_{in}=1$, $p_{out}=0$. Validation sets of size $N_i^{(val)}=500$ were stratified by cluster labels.

\begin{figure}[htbp]
	\centering
	\begin{subfigure}[b]{\linewidth}
		\centering
		\includegraphics[width=\columnwidth]{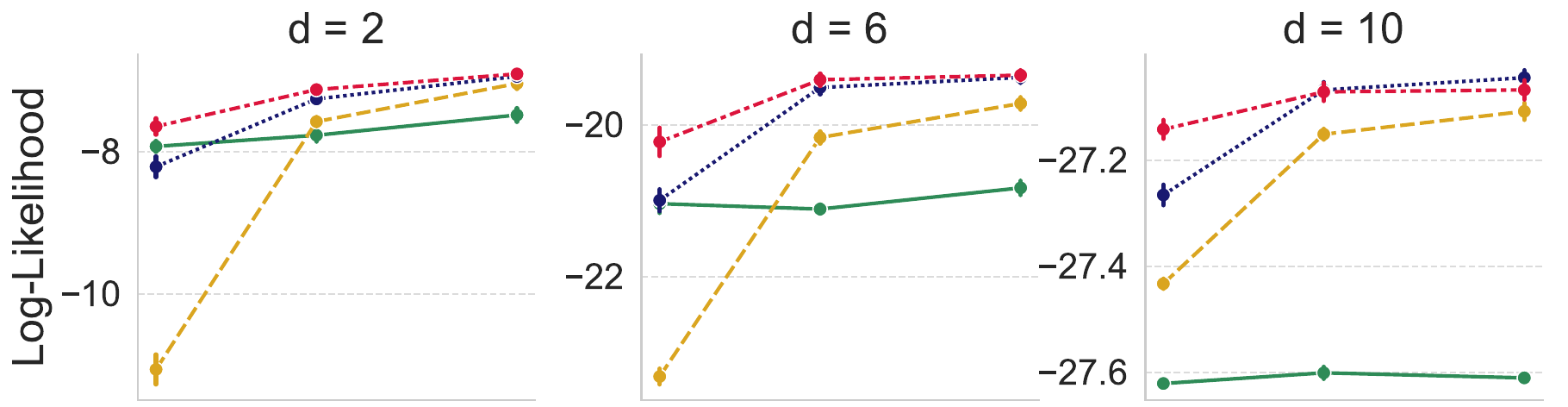}
		\label{fig:sub1}
	\end{subfigure}
	
	\begin{subfigure}[b]{\linewidth}
		\centering
		\includegraphics[width=\columnwidth]{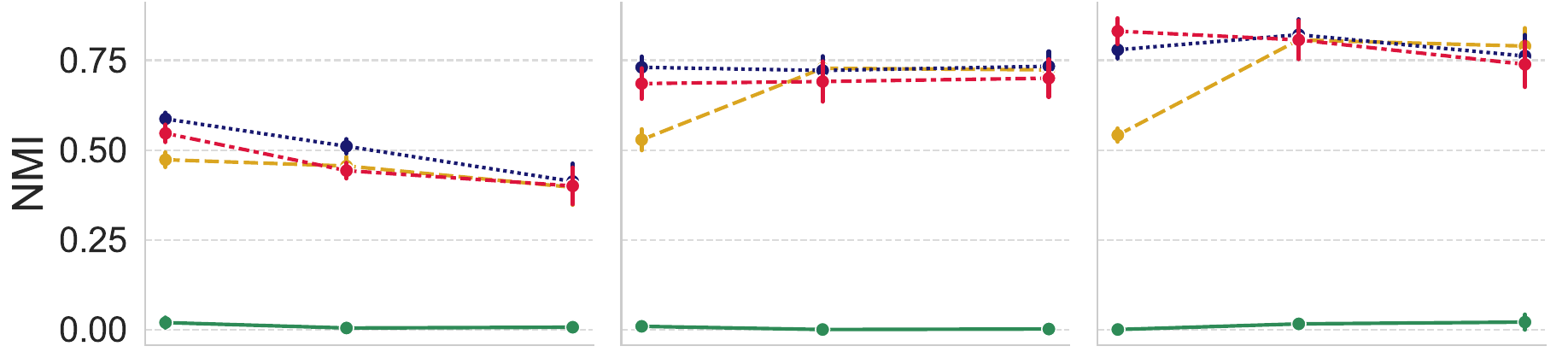}
		\label{fig:sub2}
	\end{subfigure}
	
	\begin{subfigure}[b]{\linewidth}
		\centering
		\includegraphics[width=\columnwidth]{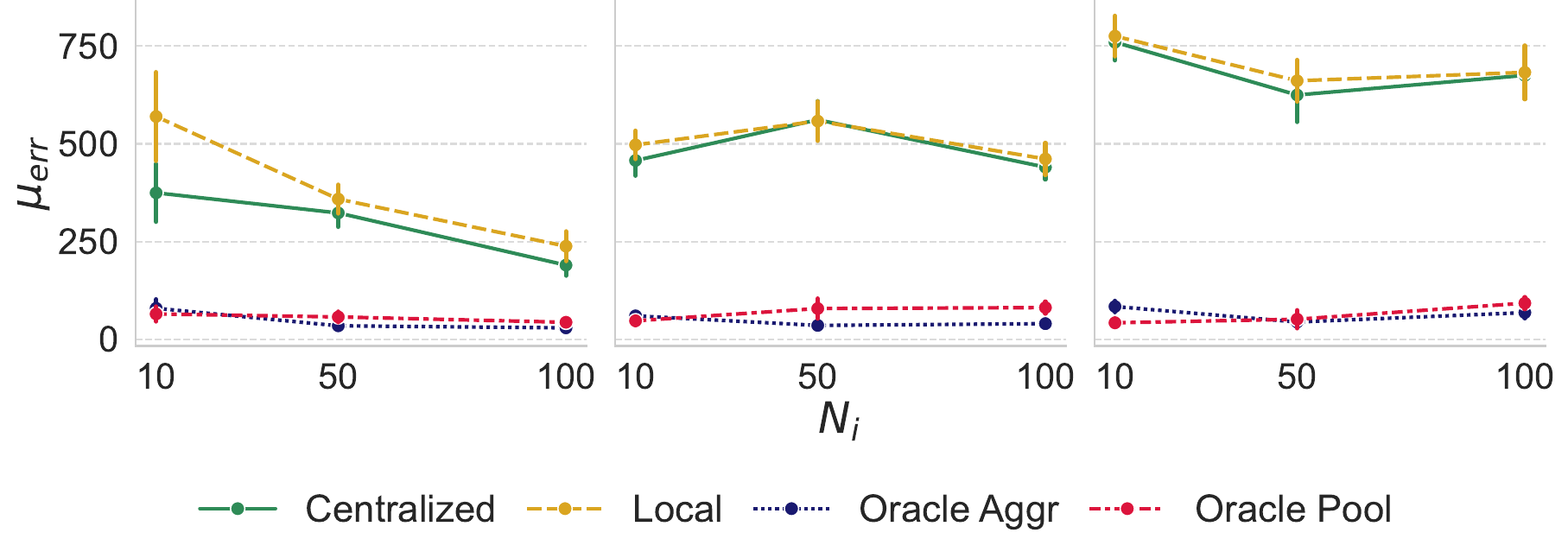}
		\label{fig:sub3}
	\end{subfigure}
	
	\caption{Comparison of methods across different dimensions. Local, centralized and GraphFed-EM training with local datasets pooled cluster-wise directly or indirectly via aggregation step. $C=5$ cluster with $N=25$ nodes, and $K=3, p_{in}=1, p_{out}=0$.}
	\label{fig:benchmarks}
\end{figure}

Results (Fig.~\ref{fig:benchmarks}) show discrepancies across metrics. Centralized EM matches Oracle log-likelihoods for $d=2,6$, but yields consistently lower NMI, likely due to overlap between cluster GMMs in low dimensions. Local training underperforms at $N_i=10$ but approaches GraphFed-EM as $N_i$ grows. Parameter error  $\mu_{err}$ is smallest for Oracles, whether pooling or aggregating.

Overall, collaboration is most beneficial when $d$ is high and $N_i$ is small. Direct pooling and GraphFed-EM with $\alpha=1$ perform similar, except at $N_i=10$, where pooling yields higher log-likelihood but no clear advantage in NMI or $\mu_{err}$.

\textbf{\textit{Cluster centroids consensus.}}
Next, we examined how aggregation strength $\alpha$ and connectivity of the graph affect consensus of the cluster centroids. We compute the weighted squared norm of the difference between the estimated mean vector $\vmu_k^i$ and the consensus of neighboring nodes $\vmu_k^{i,cons}$:

\begin{equation}
	\mu_{err}^{cons} = \frac{1}{NN_i}
	\sum_{i \in \nodes} \sum_{k=1}^{K} N_k^i \norm{\vmu_k^i - \vmu_k^{i, cons}}_2^2
	\label{eq:consensus_err}
\end{equation}
where
\begin{equation}
	\vmu_k^{i, cons} = \frac{\sum_{j \in \neib(i)} A_{ij} N_k^j \vmu_k^j}{\sum_{j \in \neib(i)} A_{ij} N_k^j }
	\label{eq:consensus}
\end{equation}
		
\begin{figure}[htbp]
	\centering
	\includegraphics[width=\columnwidth]{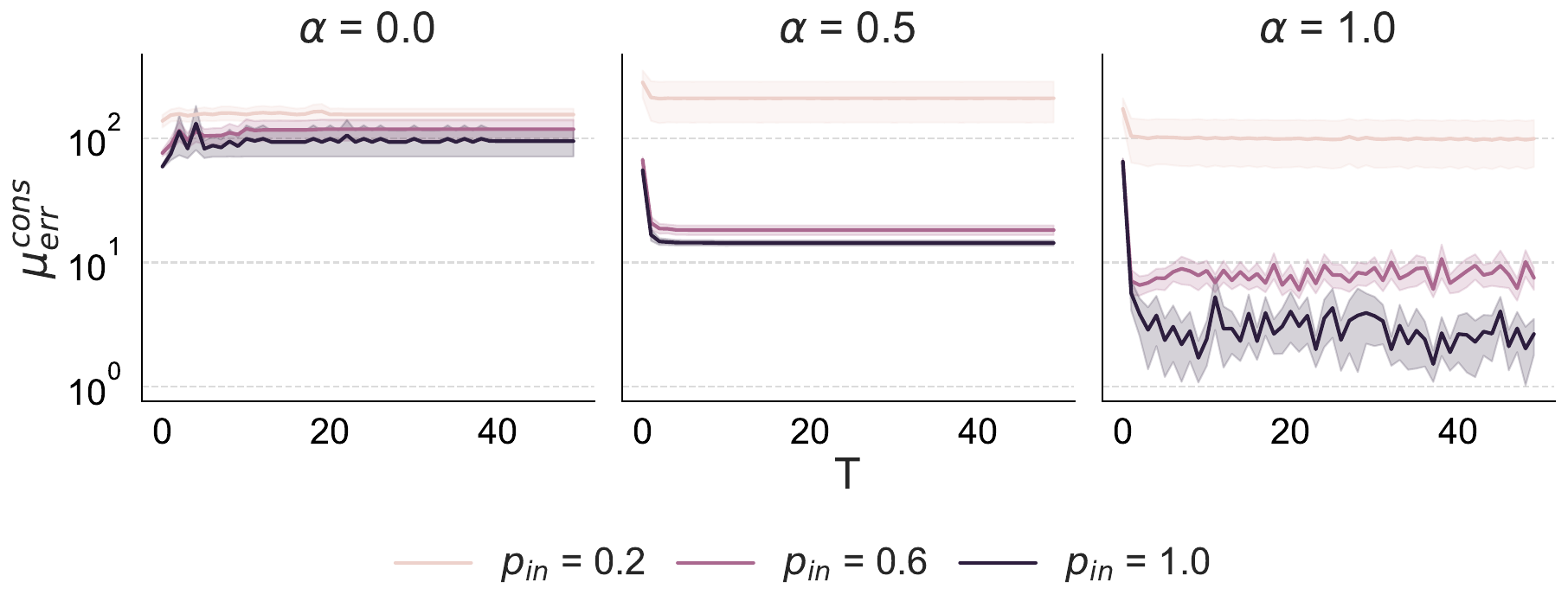}
	\caption{GraphFed-EM consensus error on one cluster with $N=10$ nodes, and $N_i=10, d=10, K=3, T=50$.}
	\label{fig:cons}
\end{figure}

Figure \ref{fig:cons} shows that increasing connectivity probability $p_{in}$ and aggregation strength $\alpha$ leads to smaller consensus error $\mu_{err}^{cons}$.

\textbf{\textit{Robustness of GraphFed-EM to out-of-cluster connections.}}
So far, we assumed the "true" connectivity matrix, where only nodes in the same cluster are connected. Next, we investigate how spurious connections between different clusters affect performance in a low sample support setting.
We test out-of-cluster connectivity probabilities $p_{out} \in \{0, 0.2, 0.4\}$ and varying aggregation strength  $\alpha$. The number of components is set to $K=3$, local sample size $N_i=10$, $d=10$, and $p_{in}=1$.

\begin{figure}[htbp]
	\centering
	\includegraphics[width=\columnwidth]{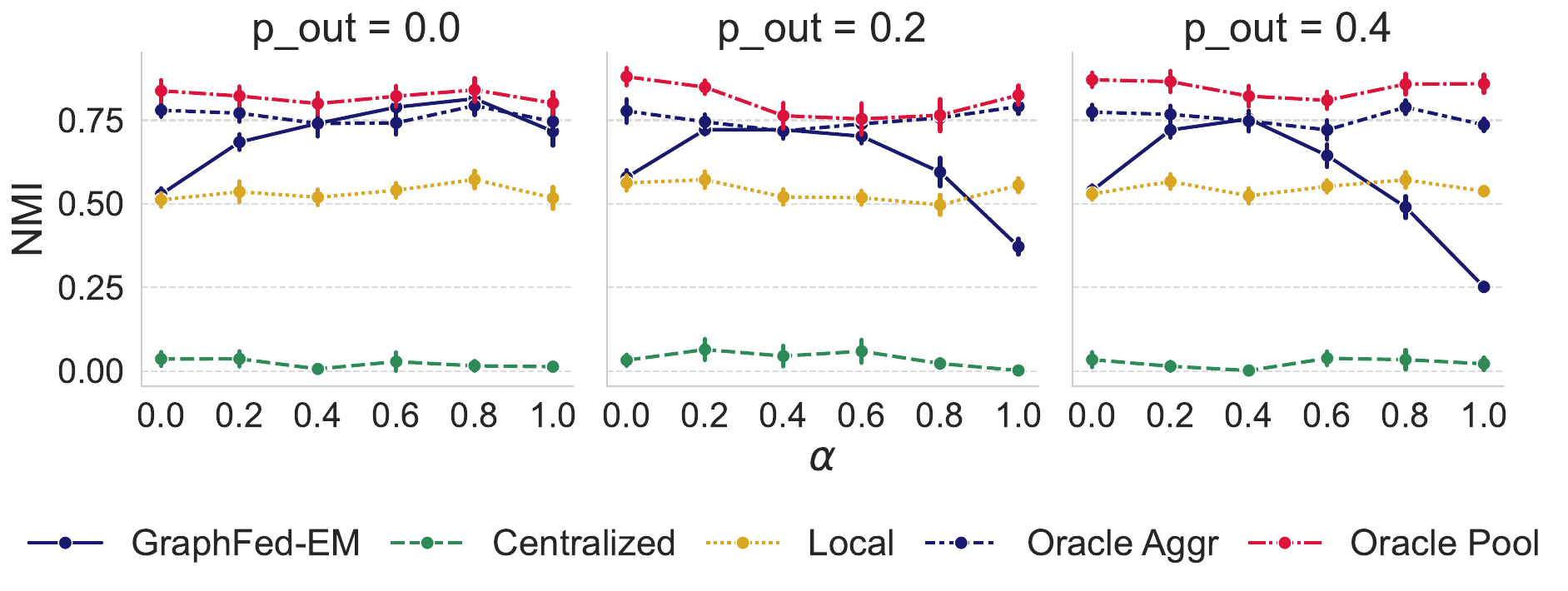}
	\caption{Effect of out-of-cluster connections $p_{out}$ and agggregation strength $\alpha$ on performance of GraphFed-EM.}
	\label{fig:alpha}
\end{figure}

As shown in Fig.~\ref{fig:alpha}, when $p_{out} = 0$, increasing $\alpha$ improves NMI. For $p_{out} > 0$, however, larger $\alpha$ leads to performance degradation. In these cases, an intermediate aggregation strength achieves the best results, matching Oracle performance (GraphFed-EM with the true connectivity matrix). Specifically, $\alpha = 0.4$ is optimal for $p_{out} \in {0.2, 0.4}$.
		
\subsection{Synthetic Dataset - client-specific priors $\pi_k^i$}
\label{subsec:het_priors}
We use a dataset with shared GMM parameters but skewed feature distributions due to different mixing coefficients $\pi_k^i$. Unlike \cite{pettersson2025federated}, we assume that this heterogeneity remains fixed during inference, and generate the validation set using the same client-specific priors $\pi_k^i$ as for training. 
We define the adjacency matrix $A$ by measuring the overlap between distributions:
\begin{align}
 A_{ij} = \frac{\sum_{k=1}^K \min(\pi_k^i, \pi_k^j)}{\sum_{j \in \neib(i)} \sum_{k=1}^K \min(\pi_k^i, \pi_k^j)}
\end{align}

We can see from Figure \ref{fig:het} that the centralized model fits the data best with higher dimensionality $d=6,10$. GraphFed-EM outperforms local, but not centralized training. Thus, in the case of client-specific priors the centralized solution is optimal for local datasets with small sample size $N_i=10$.

\subsection{MNIST -  skewed label distribution}
We construct a similar dataset with client-specific priors $\pi_k^i$ using MNIST. The dataset consists of 28x28 grayscale digit images (0 to 9). We first apply UMAP \cite{mcinnes2020umapuniformmanifoldapproximation} for dimensionality reduction to $d=\{2,6,10\}$, and set $K=10$, assuming one component per digit. Data are distributed across $N=10$ nodes with skewed lable proportions using a Dirichlet distribution (concentration parameter $\alpha=0.3$). Local sample sizes are $N_i = \{50,100,200\}$. The similarity graph is built as in Section \ref{subsec:het_priors}, based on label overlap between nodes.

\begin{figure}[htbp]
	\centering
	\begin{subfigure}[b]{\linewidth}
		\centering
		\includegraphics[width=\columnwidth]{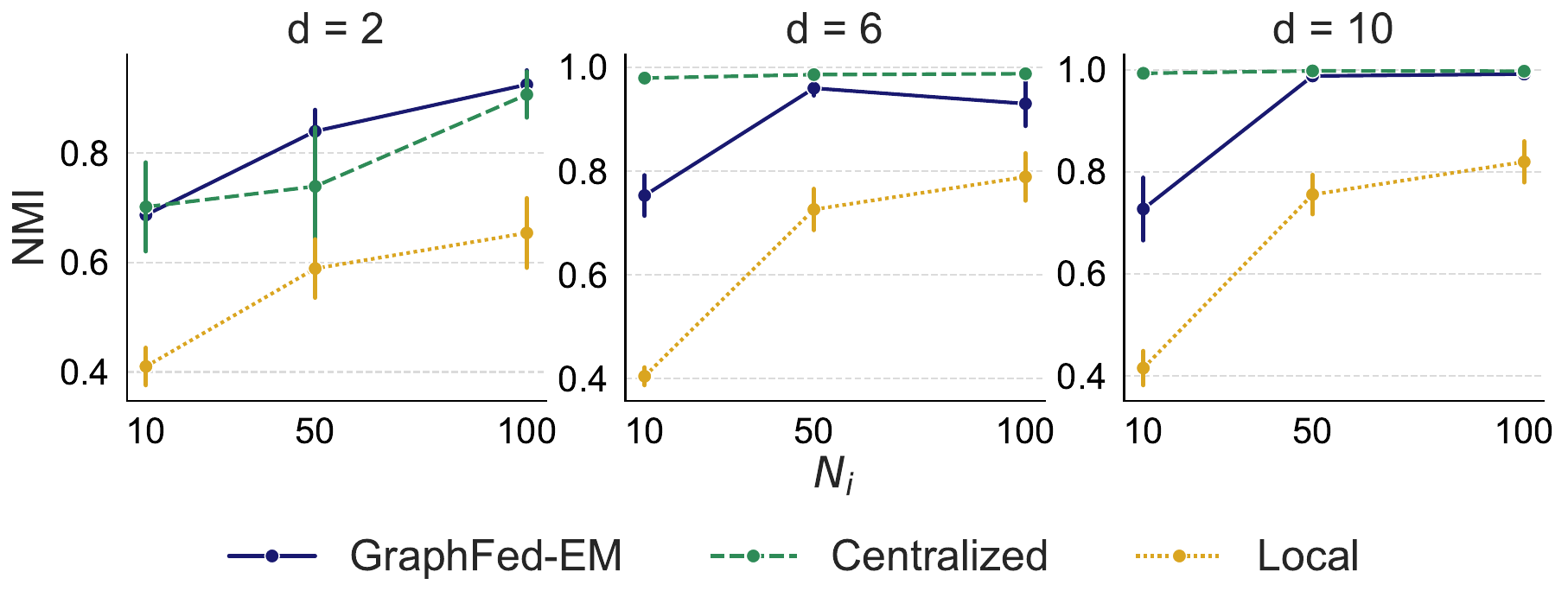}
		\caption{Performance of GraphFed-EM on dataset with client-specific priors $\pi^i$.}
		\label{fig:het}
	\end{subfigure}
	
	\begin{subfigure}[b]{\linewidth}
		\centering
		\includegraphics[width=\columnwidth]{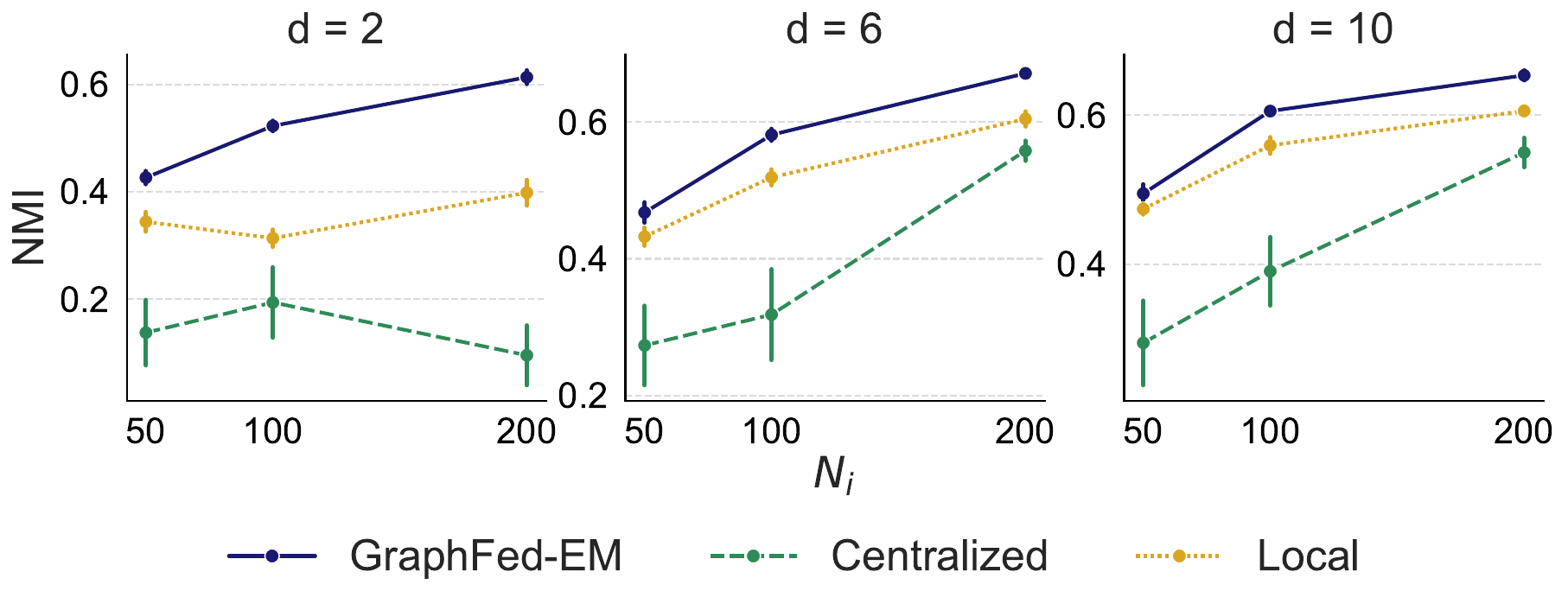}
		\caption{Performance of GraphFed-EM on MNIST dataset with skewed label distribution. $N=10, K=10$ and $T=10$.}
		\label{fig:mnist}
	\end{subfigure}
	
		
	\caption{Performance of GraphFed-EM on dataset with client-specific priors and MNIST dataset with skewed label distribution.}
	\label{fig:het_mnist}
\end{figure}

As shown in Figure \ref{fig:mnist}, GraphFed-EM achieves higher NMI values than both local and centralized GMMs across all $(d, N_i)$ pairs. 

\section{Graph-based aggregation as regularized EM}
\label{sec:analyses}
Let $\theta_k^i$ denote a generic parameter of component $k$ at node $i$ (e.g., $\vmu_k^i$, $\vcov_k^{i}$, or $\pi_k^i$), and let 
$\neib(i)$ denote the set of neighbors of node $i$ in the empirical graph.
Let $N_k^i$ be the responsibility-weighted sample count for component $k$ at node $i$. For simplicity, we assume that the GMM components of connected nodes are perfectly matched.

We can view GraphFed-EM as an instance of regularized EM, where the local EM objective is augmented with a smoothness penalty across the graph:
\begin{equation}
	\ell_i(\theta_k^i) + \frac{\lambda}{2}\sum_{j \in \neib(i)}  A_{ij} N_k^j
	\, \big\| \theta_k^i - \theta_k^j \big\|_2^2
\label{eq:pen_em}
\end{equation}
where the first term is the local negative $Q$-function (an upper bound on the log-likelihood \cite{bishop2006pattern}), $\ell_i(\theta_k^i) = -Q_i(\theta_k^i)$, and the second term penalizes dissimilarity between neighboring nodes.

If $\theta^{i, \mathrm{EM}}_k$ is the unpenalized EM maximizer of $\ell_i(\theta_k^i)$, then 
$\nabla \ell_i(\theta^{i, \mathrm{EM}}_k) = 0$.
A proximal gradient step with respect to penalty term at $\theta^{i, \mathrm{EM}}_k$ is then:
\begin{align}
	\theta_k^{i, \mathrm{new}} 
	&= \theta_k^{i, \mathrm{EM}} 
	- \eta \lambda \sum_{j \in \neib(i)} A_{ij} N_k^j
	\big(\theta_k^{i, \mathrm{EM}} - \theta_k^j \big) \nonumber \\ 
	&= (1-\alpha) \, \theta_k^{i, \mathrm{EM}} 
	+ \alpha
	\frac{N_k^i \theta_k^{i, \mathrm{EM}}  + \sum_{j} A_{ij} N_k^j \theta_k^j}{N_k^i + \sum_{j} A_{ij} N_k^j} \nonumber
\end{align}
with $\alpha = \eta\lambda (N_k^i + \sum_{j} A_{ij} N_k^j)$. This is equivalent to aggregation step Eq. \ref{eq:blending}.
It shows that the graph-based aggregation step in GraphFed-EM can be interpreted as a proximal update on a regularized EM objective: each node updates its parameter toward a weighted average of its neighbors' parameters, with weights given by the effective sample sizes $N_k^j$ and edge weight $A_{ij}$. Thus, the aggregation balances local evidence (from node $i$'s own data) with graph-based smoothing (information from neighbors), which provides a connection between GraphFed-EM and regularized EM.

\section{Conclusions}
\label{sec:conclusions}
We introduce a simple decentralized graph-regularized federated learning algorithm for Gaussian mixture models. Our approach is tailored for peer-to-peer collaborative learning and is particularly effective in heterogeneous client environments. Unlike centralized solutions, it leverages graph-based regularization to promote information sharing among related clients while respecting local data characteristics.
Numerical experiments show that GraphFed-EM effectively balances the trade-off between achieving consensus among connected nodes and optimizing local objectives. In particular, it outperforms both local and centralized training in low-sample regimes, where the data dimensionality is comparable to or larger than the number of local samples.
Our method does not require full participation of all clients and can be adapted to client-specific communication and computational constraints. Future extensions include dynamically inferring the similarity graph during communication rounds, in the spirit of clustered FL \cite{ghosh2020efficient}, and analyzing the connectivity conditions necessary to guarantee recovery of local GMM parameters and algorithmic convergence.

\section{Compliance with Ethical Standards}
This is a numerical simulation study for which no ethical approval was required.

\bibliographystyle{IEEEtran}
\bibliography{Literature}

\end{document}